\documentclass{article}

\usepackage[preprint]{neurips_2023}

\usepackage[utf8]{inputenc} 
\usepackage[T1]{fontenc}    
\usepackage{hyperref}       
\usepackage{url}            
\usepackage{booktabs}       
\usepackage{amsfonts}       
\usepackage{nicefrac}       
\usepackage{microtype}      
\usepackage{xcolor}         
\usepackage{graphicx}
\usepackage{subcaption}
\title{Minor SFT loss for LLM fine-tune to  increase performance and reduce model deviation} 

\author{%
	Shiming Xie \thanks{First author} \and Hong Chen  \and Fred Yu  \\
	\and Zeye Sun \and Xiuyu Wu  \\
	\and  \{shiming.xsm, wuyi.chen, fred.yf, zeye.szy, wuxiuyu.wxy \}@antgroup.com
}

\begin{document}

\maketitle

\begin{abstract}
Instruct LLM provide a paradigm used in large scale language model to align LLM to human preference.  The paradigm contains supervised fine tuning and reinforce learning from human feedback. This paradigm is also used in downstream scenarios to  adapt LLM to specific corpora and applications. Comparing to SFT, there are many efforts focused on RLHF and several  algorithms being proposed, such as PPO, DPO, IPO, KTO, MinorDPO and etc.  Meanwhile most efforts for SFT are focused on how to collect, filter and mix high quality data. In this article with insight from DPO and MinorDPO,  we propose a training  metric for SFT to measure the discrepancy between the optimized model and the original model, and a loss function MinorSFT that can increase the training effectiveness, and reduce the discrepancy between the optimized LLM and original LLM.	
\end{abstract}

\section{Background}
LLM trained on very large corpora is extremely powerful language model for completion tasks. SFT and RLHF(\cite{ouyang2022training}, \cite{ziegler2020finetuning}) are two techniques that used to expose the LLM capability and align LLM answer to human instructions. With the increasing reasoning abilities, LLM are widely used in industries, and SFT and RLHF are also used to inject domain knowledge into LLM by training on domain corpora. 

In the past most works are focused on RLHF and several algorithms are proposed, such as PPO{\cite{schulman2017proximal}}, DPO(\cite{rafailov2023direct}), IPO(\cite{azar2023general}), KTO(\cite{ethayarajh2024kto}), MinorDPO(\cite{xie2024minordporejectpenalty}) and etc.  One important point of RLHF is to constraint the optimized model not to deviate from the original model too much during the training, and thus PPO use KL constraints, DPO use a sample level dynamic coefficient related to distance between the preference pair, and IPO use a targeted distance between the preference pair and etc. The purpose of this constraint is to avoid over-fit on the domain corpora and to maintain LLM generalities. It's an important hypothesis that the base model is powerful enough and the training should not change the language distribution too much to maintain the generality and diversity.

While back to SFT, most works are focused on collect, filter and mix high quality data. High quality data is undoubtedly important to get a high qualified and usable LLM, while the aforementioned hypothesis that optimized model should not deviate far from the original model is still important. 

Our main contribution is that we introduce a training metrics used in DPO and MinorDPO into SFT phase, and propose an improved loss function MinorSFT. MinorSFT use a sample level coefficient to control the learning strength. It constraints the discrepancy more compared to raw SFT and may provide better performance result, at the cost of an additional hyper parameter and more computation.

\section{Related Work }
Reinforce Learning from human feedback(\cite{ouyang2022training}, \cite{ziegler2020finetuning}) is a popular technique to align LLM to human preference.   It uses SFT to train a supervised LLM on data of sampled prompt and labeled answer, then trains a reward model on preference pairs from human feedback and finally uses RL algorithm like PPO \cite{schulman2017proximal} to train an optimized LLM. The RL part contains a KL-divergence constraint to prevent the optimized LLM deviating too much from the base model.  

DPO(\cite{rafailov2023direct}) is a simplified RL algorithm that optimize LLM directly on the preference data using a cross-entropy classification loss.  DPO objective is to increase the relative log probability of preferred answer to dis-preferred answer. It incorporates a dynamic, sample level importance weight scaled by hyper-parameter $\beta$ and claim that $\beta$ account for the strength of the KL-divergence constraint. DPO introduces an important concept that LLM model itself is an implicit reward model, which means the LLM model can somehow measure the corpora during training phase. \cite{rafailov2024r} derive that DPO is token-level MDP and works as a general inverse Q-learning algorithm in a theoretical way.

MinorDPO(\cite{xie2024minordporejectpenalty}) is a DPO variant. It justifies hyper-parameter $\beta$ in DPO is a constraint relate to the relative log probability margin of the preference pair, instead of the KL-divergence constraint. It introduces MinorDPO loss to reduce penalty on the reject(dis-preferred) answer to prevent over penalty on the reject answer, which implicitly keep to the hypothesis that optimized model should not deviate too much from the base model.

IPO(\cite{azar2023general}) proves DPO may be prone to over-fitting when preferred probability over dis-preferred probability that is close to 1. In IPO objective it uses a target value relate to the hyper-parameter $\beta$ for the relative log probability of preferred to dis-preferred. However, it is somehow same as DPO, that it focuses on the relative log probability margin, so it has same problem as DPO mentioned in MinorDPO.

KTO(\cite{ethayarajh2024kto}) proposes human aware loss function. It separates the preference pair loss into two losses so that it doesn't purely rely on paired preference data. Inside each separated loss, it estimates the KL term by matching input x' with unrelated outputs z in the same batch, but without back-propagate through the KL term, and thus it also introduces an implicit constraint on the gradient which in turn affect the learning strength and deviation. 

Llama 3 (\cite{dubey2024llama3herdmodels}) presents a detailed way to collect, filter and mix high quality data for SFT and RL. For the RL part it uses DPO with an additional negative log-likelihood loss, similar to \cite{pang2024iterativereasoningpreferenceoptimization} and mentioned in \cite{pal2024smaug}. 

Many efforts focus on RL part and use explicit or implicit constraints to limit optimized LLM deviation to reduce model regression. Inspired by DPO and MinorDPO, we think it worth a try to take the hypothesis into SFT to reduce LLM deviation and maintain diversity, and maybe able to increase performance further.

\section{Approach}

\subsection{Minor SFT derivation }
DPO(\cite{rafailov2023direct}) derives its objective from RL in a closed form.

\begin{equation}
	L_{DPO}(\pi_\theta;\pi_{ref}) 
	=-\mathbb E_{(x,y_w,y_l) \sim D}[{log\sigma(\beta log \frac{\pi_\theta(y_w|x)}{\pi_{ref}(y_w|x)} - \beta  log \frac{\pi_\theta(y_l|x)}{\pi_{ref}(y_l|x)})} ] \label{dpo_equation}
\end{equation}

It introduces $\hat{r}_\theta(x,y) = \beta log \frac{\pi_\theta(y|x)}{\pi_{ref}(y|x)} $ as the reward implicitly defined by the language model $\pi_\theta$ and reference model $\pi_{ref}$.  DPO objective is to maximize rewards margin between the preference pair.

The MinorDPO objective adds an additional constraints to dis-preferred samples, by replacing the original penalty $log\frac{\pi_\theta(y_l|x)}{\pi_{ref}(y_l|x)}$ with $max(0, log\frac{\pi_\theta(y_l|x)}{\pi_{ref}(y_l|x)})$. 

\begin{equation}
	L_{MinorDPO}(\pi_\theta;\pi_{ref}) \\
	=-E_{(x,y_w,y_l) \sim D} {log\sigma(\beta log \frac{\pi_\theta(y_w|x)}{\pi_{ref}(y_w|x)} - \beta max(0, log \frac{\pi_\theta(y_l|x)}{\pi_{ref}(y_l|x)}))}   \label{minor_dpo_equation}
\end{equation}

So when probability of optimized model on dis-preferred sample is less than probability of reference model on dis-preferred sample,  which means $ log\frac{\pi_\theta(y_l|x)}{\pi_{ref}(y_l|x)} <= 0$ and so $max(0, log\frac{\pi_\theta(y_l|x)}{\pi_{ref}(y_l|x)}) = 0$, it will ignore the dis-preferred , and  focus on handling the preferred sample. 

Under this situation, the formula can be rewritten as below:

\begin{equation}
	L_{Preferred}(\pi_\theta;\pi_{ref}) \\
	=-E_{(x,y) \sim D} {log\sigma(\beta log \frac{\pi_\theta(y|x)}{\pi_{ref}(y|x)})}  \label{preferred_equation}
\end{equation}

We simply name this method SFT using DPO. Eq. \ref{preferred_equation} tries to maximize reward on the preferred sample. 

Let's derive the gradient equation of Eq. \ref{preferred_equation} 

\begin{equation}
	\nabla_\theta L_{preferred}(\pi_\theta;\pi_{ref}) \\
	=-\mathbb E_{(x,y) \sim D}[\beta\sigma(-\beta log \frac{\pi_\theta(y|x)}{\pi_{ref}(y|x)} )[\nabla_\theta log\pi_\theta(y|x) ] ] \label{preferred_equation_gradient}
\end{equation}

Compared to raw SFT loss gradient equation.

\begin{equation}
	\nabla_\theta L_{raw\_sft}(\pi_\theta;\pi_{ref}) \\
	=- \mathbb E_{(x,y) \sim D}[\frac{1}{m} \nabla_\theta log\pi_\theta(y|x) ]  \label{raw_sft_gradient}
\end{equation}
$m$ is length of the answer. Normally, SFT use average over the answer, while DPO use sum over the answer.

Comparing Eq. \ref{preferred_equation_gradient} and Eq. \ref{raw_sft_gradient}, we see Eq. \ref{preferred_equation_gradient} $\nabla_\theta L_{preferred}$ has three part: a hyper-parameter $\beta$, a sample level dynamic coefficient $\sigma(-\beta log \frac{\pi_\theta(y|x)}{\pi_{ref}(y|x)})$, and a sample related gradient  $\nabla_\theta log\pi_\theta(y|x) $. 

While Eq. \ref{raw_sft_gradient}  $\nabla_\theta L_{raw\_sft}$ contains two parts: a sample answer  related length $\frac{1}{m}$ and a sample relate gradient  $\nabla_\theta log\pi_\theta(y|x) $.

Here we introduce the sample level dynamic coefficient $\sigma(-\beta log \frac{\pi_\theta(y|x)}{\pi_{ref}(y|x)})$ into raw sft loss, and we get

\begin{equation}
	\nabla_\theta L_{minor\_sft\_naive}(\pi_\theta;\pi_{ref}) \\
	=- \mathbb E_{(x,y) \sim D}[\frac{1}{m} \sigma(-\beta log \frac{\pi_\theta(y|x)}{\pi_{ref}(y|x)}) \nabla_\theta log\pi_\theta(y|x) ]  \label{minor_sft_gradient_naive}
\end{equation}

Since at the start of the training, $\pi_\theta$ is same as $\pi_{ref}$, so $\sigma(-\beta log \frac{\pi_\theta(y|x)}{\pi_{ref}(y|x)}) ==\sigma(0) == 0.5 $, so we multiply 2 to make it closer to the raw sft and  get final MinorSFT gradient.

\begin{equation}
	\nabla_\theta L_{MinorSFT}(\pi_\theta;\pi_{ref}) \\
	=- \mathbb E_{(x,y) \sim D}[\frac{2}{m} \sigma(-\beta log \frac{\pi_\theta(y|x)}{\pi_{ref}(y|x)}) \nabla_\theta log\pi_\theta(y|x) ]  \label{minor_sft_gradient}
\end{equation}

\subsection{LLM Deviation metric}
Back to the reward aforementioned $\hat{r}_\theta(x,y) = \beta log \frac{\pi_\theta(y|x)}{\pi_{ref}(y|x)} $, DPO objective is to maximize the rewards margin between the preference pairs. And we can also treat the reward as a metric that measure
\begin{enumerate}
	\item complexity of the sample. As the reward is $ \beta (log \pi_\theta(y|x) - log \pi_{ref}(y|x)) $, high rewards mean $\pi_\theta$ gives high log probability, which indicate the sample is low complexity. 
	\item deviation of the model. If we treat the corpora as identical distribution, then high rewards mean high relative log probability difference between the optimized LLM $\pi_\theta$ and the original LLM $\pi_{ref}$, which indicate a high deviation.
\end{enumerate}

So the sample dynamic coefficient $\sigma(-\beta log \frac{\pi_\theta(y|x)}{\pi_{ref}(y|x)})$ in Minor SFT has  a clear physic meaning,\textbf{ lower complexity samples have a smaller coefficient than higher complexity samples}. In this way it dynamically adjusts the training data distribution and the whole training process will pay more attention on higher complexity samples.

Besides, this metric measures how far the optimized model deviate from the original model during training. But it has two limitations: 

\begin{enumerate}
	\item The reward is related to the hyper-parameter $\beta$, so rewards of different $\beta$ is not able to do comparison. 
	\item This reward is related to answer length. so training of distribution with different answer length is not able to do comparison. 
\end{enumerate}

We need a normalized metric that can be compared not only with different $\beta$, but also with corpora of different answer length. so here we need to normalized both $\beta$ and answer length, and get
\begin{equation}
	m_\theta(x,y) = \frac{1}{N} \Sigma \frac{1}{m} log \frac{\pi_\theta(y|x)}{\pi_{ref}(y|x)} 
\end{equation}
N is batch size, and m is answer length. Thus $m_\theta(x,y) $ can somehow be used as a training metric to measure the deviation between the optimized model $\pi_\theta$ and the reference model $\pi_{ref}$. Even SFT do not use hyper-parameter $\beta$, it can compare with MinorSFT and SFT using DPO( Eq. \ref{preferred_equation})

\begin{figure}[htbp]
	\centering
	\begin{minipage}{0.32\textwidth}
		\includegraphics[width=\textwidth]{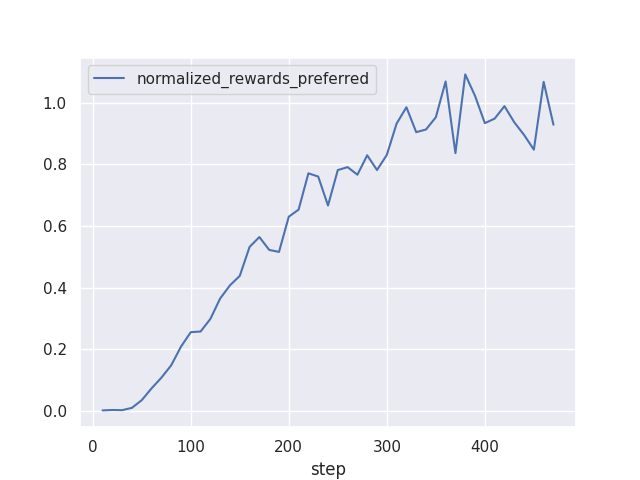} 
		\subcaption{raw sft lr = 1e-5}	\label{normalized_reward_compare_1em5_sft}	
	\end{minipage}
	\begin{minipage}{0.32\textwidth}
		\includegraphics[width=\textwidth]{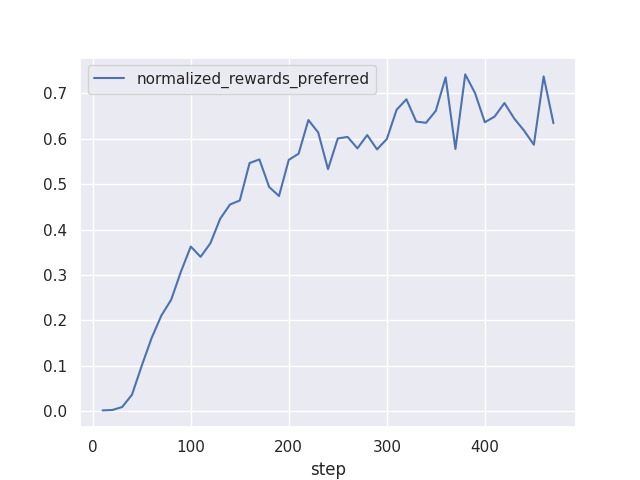} 
		\subcaption{SFT use DPO lr = 2e-5 $\beta$ = 0.04} 	\label{normalized_reward_compare_1em5_sft_use_dpo}	
	\end{minipage}
	\begin{minipage}{0.32\textwidth}
		\includegraphics[width=\textwidth]{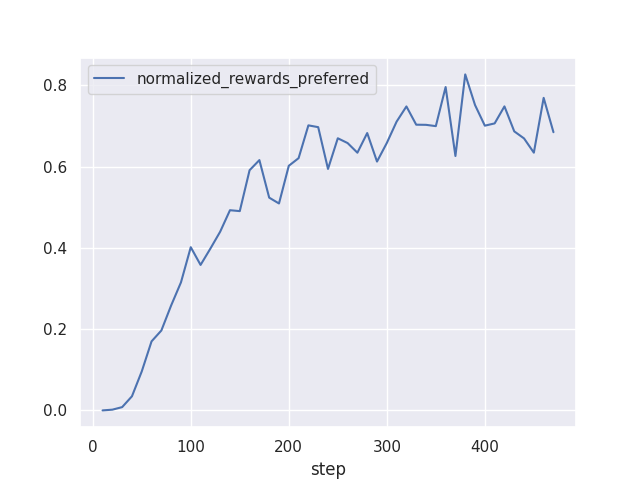} 
		\subcaption{Minor SFT lr = 2e-5 $\beta$ = 0.04} 	\label{normalized_reward_compare_1em5_minor_sft}	
	\end{minipage}
	
	\caption{Normalized rewards during training}
	\label{normalized_reward_compare_1em5}	
\end{figure}

The metric $m_\theta(x,y) $ can be used in both DPO with preference pair and SFT with only preferred. Figure \ref{normalized_reward_compare_1em5} shows metric trends for three methods. Since the optimized LLM deviate from the reference model during to the training, they can also be treated as LLM model deviation trend. The metric value and trends can be used in a qualitative analysis of LLM deviation.  And from Figure \ref{normalized_reward_compare_1em5} even with larger learning rate(2e-5) for  SFT use DPO \ref{normalized_reward_compare_1em5_sft_use_dpo} and MinorSFT \ref{normalized_reward_compare_1em5_minor_sft}, they have a lower deviation value compared to SFT (1e-5) \ref{normalized_reward_compare_1em5_sft} for each training step, due to they both have a sample level dynamic coefficient that decays fast when the reward of the sample grow up( or in other words, when the complexity of the sample reduce down).

\section{Experiments}
For training settings, we use Qwen2-7B-Instruction(\cite{qwen2}) as the base model. It expresses high performance in many benchmarks \footnote{We tried several open datasets to train the base model, but with little performance improvement on the benchmarks, so in this experiment we use a private domain corpus}. And use down-sample of FinanceIQ\footnote{https://huggingface.co/datasets/Duxiaoman-DI/FinanceIQ}, fineval\footnote{https://huggingface.co/datasets/djdropthebit/fineval}, ceval-exam(\cite{huang2023ceval}) as test datasets to do evaluation.

We use LLaMa-Factory(\cite{zheng2024llamafactory}) as the training and inference framework with some customized code to implement the MinorSFT and SFT use DPO algorithm. The experiments use batch size 64, warm-up ratio 0.1, linear decay learning rate, 1 epoch and run 400+ steps.

For FinanceIQ and fineval we use the prompt """Please answer the questions based on the context provided. Please ensure that the original information (such as numbers, time, entities, opinions, etc.) is accurately cited when answering.
If the user's question cannot be answered based on the given context, please briefly explain why.
If the answer involves mathematical calculations, please give priority to calling tools; if it involves numerical comparison, please give the comparison process; if it involves analysis or reasoning, please give the reasoning and analysis process""". 

For ceval-exam we use the prompt """You need to choose one of the four options A, B, C, and D as the most appropriate answer to the question. You can only output one character, and this character must be one of A, B, C, and D.
The question content is: <question>
The four options are:
A. <A>
B. <B>
C. <C>
D. <D>
Your answer is:""".

\begin{figure}[htbp]
	\centering
	\begin{minipage}{0.45\textwidth}
		\includegraphics[width=\textwidth]{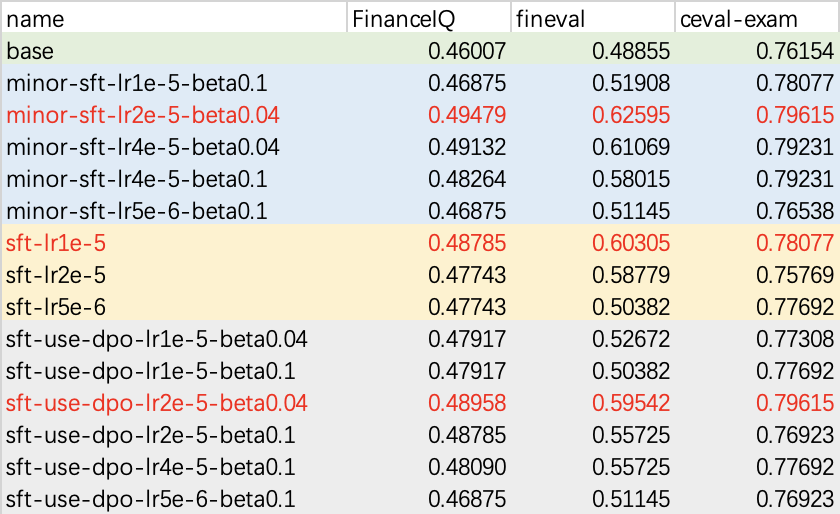} 
		\subcaption{Full comparison data }
		\label{accuracy_comparison_full_data}
	\end{minipage}
	\begin{minipage}{0.45\textwidth}
		\includegraphics[width=\textwidth]{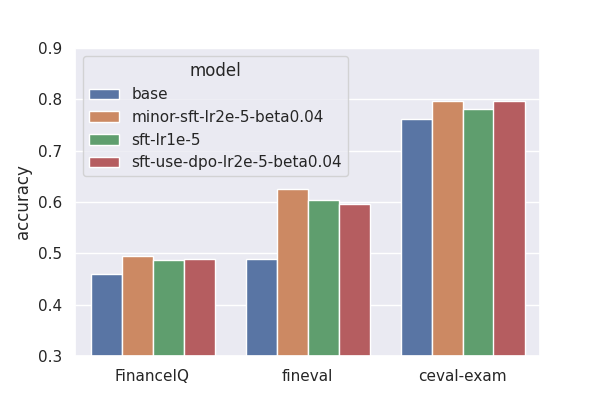} 
		\subcaption{Best model comparison}
		\label{accuracy_comparison_best}
	\end{minipage}
	\caption{Accuracy comparison}
	\label{accuracy_comparison}	
\end{figure}

Figure \ref{accuracy_comparison} shows the experiment result. We searched a group setting for each method. Figure \ref{accuracy_comparison_full_data} contains full detail of the comparison. Figure \ref{accuracy_comparison_best} shows the comparison between the best result of each method. 

Figure \ref{accuracy_comparison_full_data} shows after learning, Minor SFT get its best result with lr=2e-5 and $\beta$=0.04, raw SFT get its best result with lr=1e-5, and SFT use DPO get its best result with lr=2e-5 and $\beta$=0.04.

Figure \ref{accuracy_comparison_best} indicates that Minor SFT(lr=2e-5, $\beta$=0.04) , raw SFT(lr=1e-5), SFT use DPO(lr=2e-5, $\beta$=0.04) are all better than the base model. \textbf{Minor SFT perform best in all three datasets compared to raw SFT and SFT use DPO}. SFT use DPO wins FinanceIQ and ceval-exam but lose fineval compared to raw SFT.

The experiment result shows several points.

\begin{enumerate}
	\item Each method have a performance increase from low learning rate to high learning rate, and get a performance decrease if continue to increase the learning rate after a certain threshold. Raw SFT get its best at lr=1e-5, MinorSFT and SFT use DPO get its best at lr=2e-5 and $\beta$=0.04.
	\item Minor SFT perform best in all three datasets. We give credit to the sample-level dynamic coefficient $\sigma(-\beta log \frac{\pi_\theta(y|x)}{\pi_{ref}(y|x)})$. \textbf{This coefficient implicitly adjust the corpus distribution, so that the training pays more effort on those high complexity(or difficult) samples}.  
	\item Minor SFT need higher learning rate to get its best performance compared to raw SFT, because the sample dynamic coefficient $\sigma(-\beta log \frac{\pi_\theta(y|x)}{\pi_{ref}(y|x)})$ decay during training when the reward grows up(or when the complexity of the sample reduce down). However, even with high learning rate Minor SFT has a lower deviation compared to raw SFT, which can be see through Figure \ref{normalized_reward_compare_1em5}. 
	\item SFT use DPO perform worse than Minor SFT, we think the cause is due to it use the same hyper-parameter $\beta$ for all samples. $\beta$ is somehow used as an average factor same as $\frac{1}{m}$ in raw SFT since DPO use sum over the answer.  $\frac{1}{m}$  is sample dependent while $\beta$ is sample independent, this bias cause the performance regression.
	\item $\beta$ has same meaning as in DPO, however it still brings more complexity compared to raw SFT. It needs some tuning to achieve the best performance.
	
\end{enumerate}

\section{Conclusion \& Future work}
Inspired from DPO and MinorDPO, in this article we propose a training metric $m_\theta(x,y)$ that can used to analysis LLM deviation for SFT phase, and we propose Minor SFT that introduce an dynamic sample level coefficient $\sigma(-\beta log \frac{\pi_\theta(y|x)}{\pi_{ref}(y|x)})$ that implicitly adjust corpora distribution and prevent optimized LLM deviate from the reference model too much. Minor SFT can be used in LLM preference alignment or downstream task fine-tuning to get better performance and reduce deviation. However, due to the coefficient, MinorSFT introduce additional computation cost from the reference model and additional complexity from the hyper-parameter $\beta$. It's kind of a tradeoff to get better performance.

As the conclusion in above Experiment section, MinorSFT needs some higher learning rate compared to raw SFT. We design the MinorSFT coefficient same as the coefficient in DPO to simplify its meaning and understanding.  The hyper-parameter $\beta$ in MinorSFT has same meaning as in DPO.  With appropriate tuning we are able to get a best performance LLM as in above experiment.  

Although the training metric $m_\theta(x,y)$ can somehow be used to analysis how far the optimized LLM deviate from the reference model for different $\beta$ and  answer length, we don't have a way to know whether the optimized model is over-fit or under-fit during the training. It needs more research effort to find those metrics that can guide model's fitting level.

\bibliography{main}

\end{document}